\def\year{2021}\relax
\documentclass[letterpaper]{article} %
\usepackage{aaai21}  %
\usepackage{times}  %
\usepackage{helvet} %
\usepackage{courier}  %
\usepackage[hyphens]{url}  %
\usepackage{graphicx} %
\usepackage[ruled,vlined,linesnumbered]{algorithm2e}
\usepackage{natbib}  %
\usepackage{caption} %
\usepackage{amsmath}
\usepackage{mathtools}
\usepackage{amsfonts}
\usepackage{amssymb}
\usepackage{bbm}
\newcommand{\vs}{x}
\newcommand{\vu}{u}
\newcommand{\vg}{g}

\newcommand{\vo}{o}
\newcommand{\vp}{p}

\newcommand{\sstate}{\mathcal{X}}
\newcommand{\ssfree}{\mathcal{F}_t}
\newcommand{\ssgoal}{\mathcal{X}_\mathrm{goal}}

\newcommand{\ji}{^{ij}}

\def\HiLi{\leavevmode\rlap{\hbox to \hsize{\color{yellow!50}\leaders\hrule height .8\baselineskip depth .5ex\hfill}}}

\usepackage{tikz}
\usetikzlibrary{fit,calc}
\usepackage{xcolor}

\title{Joint-Space Multi-Robot Motion Planning with Learned Decentralized Heuristics}

\author{
    Fengze Xie,
    Marcus Dominguez-Kuhne,
    Benjamin Riviere,
    Jialin Song,
    Wolfgang Hönig,\\
    Soon-Jo Chung,
    Yisong Yue,
}
\affiliations{
    California Institute of Technology \\
    \{fxxie, mddoming, briviere, jssong, whoenig, sjchung, yyue\}@caltech.edu    
}

\begin{document}

\maketitle

\begin{abstract}

In this paper, we present a method of multi-robot motion planning by biasing centralized, sampling-based tree search with decentralized, data-driven steer and distance heuristics. 
Over a range of robot and obstacle densities, we evaluate the plain Rapidly-expanding Random Trees (RRT), and variants of our method for double integrator dynamics. 
We show that whereas plain RRT fails in every instance to plan for $4$ robots, our method can plan for up to 16 robots, corresponding to searching through a very large 65-dimensional space, which validates the effectiveness of data-driven heuristics at combating exponential search space growth. 
We also find that the heuristic information is complementary; using both heuristics produces search trees with lower failure rates, nodes, and path costs when compared to using each in isolation. 
These results illustrate the effective decomposition of high-dimensional joint-space motion planning problems into local problems.

\end{abstract}

\section{Introduction}
\label{sec:introduction}

Applications for autonomous teams of robots are rapidly expanding into domains of urban search and rescue, space and sea exploration, self-driving vehicles, and warehouse robotics. However, a critical mid-level component of the autonomy hierarchy, multi-robot motion planning, remains an active area of research because of the non-convexity of the underlying optimization problem and the exponential growth of the search space with the number of robots.

We are in particular interested in kinodynamic motion-planning, where robots obey dynamic constraints in a continuous state space; as opposed to path-planning, which usually ignores dynamics and plans in discrete space. A common kinodynamic motion planning technique is to use random sampling to grow a search tree, e.g., Rapidly-Exploring Random Tree (RRT)~\cite{RRT-base} and Expansive Space Tree (EST)~\cite{EST}. Theoretical analysis of the RRT algorithm has shown it to be probabilistically complete~\cite{kinodynamicRRT} and variants have been proposed with asymptotic optimality properties~\cite{Karaman_2011,AO-RRT}. However, sampling-based methods do not perform well for systems with complex dynamics or the large search dimension of robot teams~\cite{Lavalle_2006}. Other classical multi-robot methods include~\cite{Zhou_2017,Ames_2017,ORCA} but these do not use a tree-structure and cannot guarantee the completeness property of the algorithm, i.e., if a feasible solution exists, the algorithm will find it. 

\begin{figure} %
	\centering
    \includegraphics[width=0.95\linewidth]{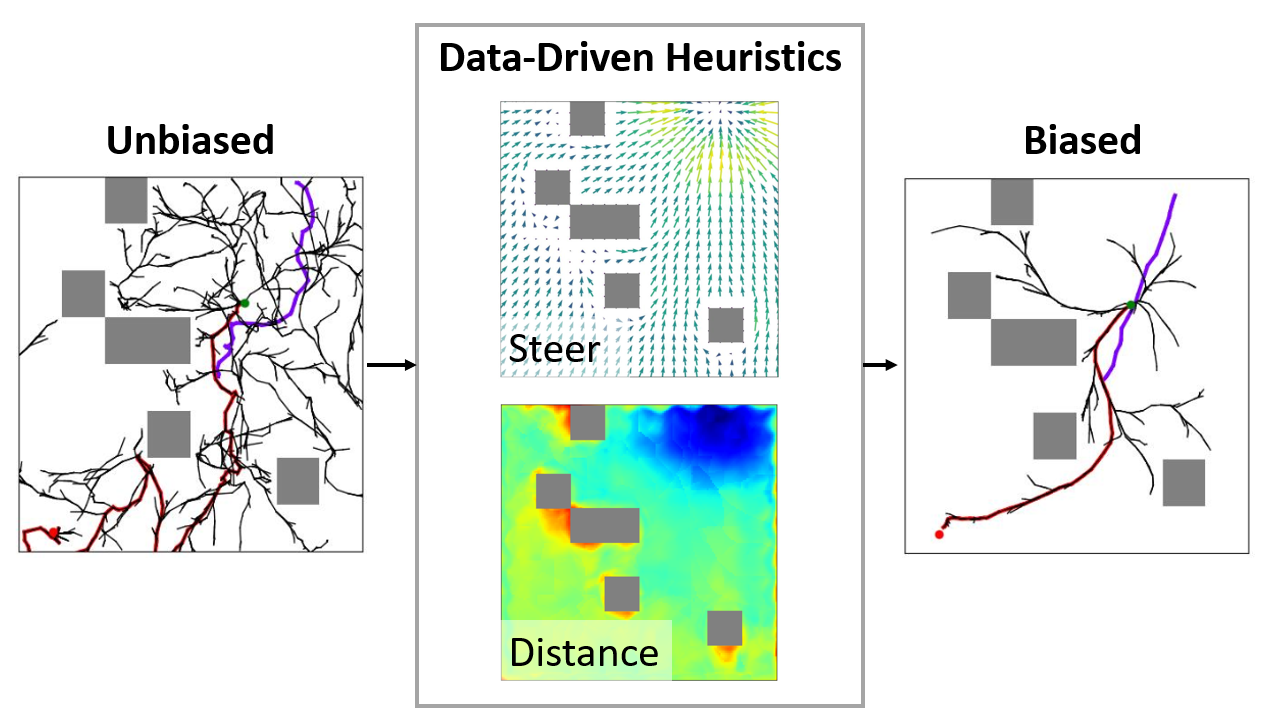}
    \caption{Schematic of our approach: Learned steer and distance heuristics guide the tree search, resulting in effective high-dimensional search and solutions with lower path cost, lower wall-clock time, fewer nodes, and lower failure rate.}
    \label{fig:overview}
        \vskip -0.1in
\end{figure}

Recently, machine learning techniques have been applied to robotic planning. Some single-agent methods tackle search complexity by biasing the tree search with data-driven heuristics~\cite{Silver_2017,DBLP:conf/icra/IchterHP18,RL-RRT,Chen_2020}. Our work extends this approach to multi-robot planning with decentralized heuristics compatible with variable team sizes. As decentralized heuristics compute control inputs for each robot with local information, the performance of these heuristics in a joint-space context illustrates the effective decomposition of high-dimensional joint-space motion planning problem into local motion planning problems. 

Existing data-driven methods either use reinforcement learning or imitation learning paradigms. Reinforcement learning methods~\citep{RL-RRT} are very general, but typically suffer from long training time. 
Instead, we use imitation learning from a centralized expert to effectively learn both heuristics with permutation-invariant neural network  models based on Deep Sets \citep{Zaheer_2017}. Our network encoding also permits a continuous state representation for dynamically coupled motion planning. Other multi-agent planners are discrete~\citep{PRIMAL} and cannot handle dynamical constraints necessary for certain robotic applications. \citet{GLAS} also uses Deep Sets to study continuous multi-robot planning and learns the steering function from data but, without integrating the policy into a tree structure, it lacks interpretability and completeness guarantees. In this work, we address this limitation by integrating data-driven distance and steering heuristics into RRT and inheriting its probabilistic completeness property. 

The overview of our method is as follows. First, we train a steer function that controls a single robot to a goal state using trajectory data from a centralized planner with a global-to-local imitation learning method. Next, we train a distance function with cost-to-go information from rolling out trajectories with the steer function. Then, both decentralized heuristics are integrated into a centralized RRT planner for joint-space planning. 
Our decentralized heuristics are compatible with arbitrary team sizes and enjoy computational efficiency and performance advantage in high-dimensional joint-space. Integrating them with a centralized planner leads to improved interpretability, probabilistic completeness guarantee, and a higher-level safety guarantee.
Both heuristics and their effect on RRT are visualized in Fig.~\ref{fig:overview}. To the best of our knowledge, this work is the first to use decentralized heuristics to guide centralized multi-robot search in a continuous state with dynamical constraints. The success of our method demonstrates the effective decomposition of high-dimensional joint-space motion planning problems into local problems. 

We empirically evaluate our method on a range of obstacle and robot density motion planning problems and show that equipping traditional planners with our heuristics is necessary to achieve a non-zero success rate for $4$ robots, and permits planning for up to $16$ robots. We also show that steer and distance heuristics each benefit the search independently by comparing variants of our method.

\section{Approach}
\label{sec:problem_formulation}

\textit{Problem Statement}: We consider the classical kinodynamic motion planning problem~\cite{AO-RRT}: given a state space, $\sstate$, a feasible control space, $\mathcal{U}$, a start state, $\vs_0 \in \sstate$, kinematic feasible set, $\ssfree \subset \sstate$, dynamical constraints $\vs_{t+1} = f(\vs_t,\vu_t)$, and goal set, $\ssgoal \subset \sstate$, find a series of admissible control inputs $\{\vu_t\}_{t \in [0,T]}, \ \vu_t \in \mathcal{U}$, that result in a motion that starts at the initial condition, ends in the goal set, and remains in the free space for the entire trajectory, i.e., $\vs_0=\vs_0$ , $\vs_t \in \ssfree$, $\vs_T \in \ssgoal$ where $T$ is the last timestep in the motion plan. 

The start state, $\vs_0$ is constructed from composing the initial states of each robot, $\vs^i_0$, e.g., $\vs_0 = [\vs^1_0; \hdots; \vs^{|\mathcal{V}|}_0]$, where the set of all robot indices is denoted $\mathcal{V}$, the robot index is denoted with a superscript, and the time index is denoted with a subscript. 
This joint-space representation couples the dimension of the state space, $\sstate$, to the number of robots. 
In particular, we consider the double integrator system in a two-dimensional space where, for robot $i$ at time $t$, the state, control, and dynamics are defined as:
\begin{align}
    \vs^i_t = 
    \begin{bmatrix}
        p^i_t \\ v^i_t \\ t 
    \end{bmatrix}, \quad
    \vu^i_t = a^i_t, \quad 
    \vs^i_{t+\Delta_t} = 
    \begin{bmatrix}
        p^i_t \\ v^i_t \\ t 
    \end{bmatrix} + 
    \begin{bmatrix}
        v^i_t \\ a^i_t \\ 1
    \end{bmatrix} \Delta_t,
\end{align}
where $p,v,a$ are position, velocity, and acceleration vectors in $\mathbbm{R}^2$, $\Delta_t$ is the simulation timestep and $t$ is the current time. The dynamics correspond to the Propagate function in line~\ref{alg:propagate} of Algorithm~\ref{algo:rrt}. The kinematic feasibility space $\ssfree$ imposes two constraints: upper bound on velocity and collision avoidance with other agents and obstacles: 
\begin{align}
    \ssfree = \{ \vs_t \ | \ 
    &\|v^i_t\|_2 \leq \overline{v}, \ \ 
    \|p^i_t - p^j_t\|_2 \geq 2 r_{\text{robot}}, \nonumber \\ 
    & p^i_t \not \in \Omega, \ \forall i,j \in\mathcal{V} ,j\neq i\}
\end{align}
where $\Omega$ is the set union of all obstacle positions and $r_\text{robot}$ is the robot radius. The admissible control space, $\mathcal{U}$, imposes a bounded control input: $\mathcal{U} = \{ u \ | \ \|u^i_t\|_2 \leq \overline{u}, \ \forall i,t\}$. Finally, the goal set $\ssgoal$ is defined as the ball of radius $r_\text{goal}$ around each robot's goal state, $\vg^i$:
\begin{align}
    \ssgoal &= \{ \vs \ | \ \sum_{i\in\mathcal{V}} \| \vs^i - \vg^i \|^2_2 \leq r_\text{goal} \}.
\end{align}

To empirically evaluate the proposed method, we consider the following metrics: success(/fail) rate, number of nodes, and path cost. Each motion planning problem is counted as a success unless the wall-clock time exceeds a time-out threshold value before reaching the goal set. The number of nodes is the size of the tree at problem termination. The path cost is calculated in an optimal control sense by integrating the norm of the control input across the entire trajectory:
\begin{align}
    \label{eq:path_cost}
    c &= \sum_i \sum_t \|u^i_t\|^2_2 \Delta_t.
\end{align}

\textit{Method}: 
Our approach is to augment high-dimensional kinodynamic RRT with decentralized steer and distance heuristics, see Algorithm~\ref{algo:rrt} where our changes from the traditional algorithm are on Lines \ref{alg:distStart} -- \ref{alg:distEnd} and \ref{alg:steerStart} -- \ref{alg:steerEnd}. 
First, we recap RRT, then we introduce the heuristics and their integration into Algorithm~\ref{algo:rrt}, and finally we discuss how they are trained. 

\begin{algorithm}%
\SetInd{1ex}{1ex}
\SetKwInOut{Input}{Input}
\caption{RRT With Decentralized Heuristics}
\label{algo:rrt}
\Input{$x_0, \ssgoal, \ssfree$}
 $\mathcal{T} = \text{Tree}(x_0)$ \;
 \While{$\text{True}$}{
    $x_\text{rand} \leftarrow \mathbbm{U}(\sstate)$ \label{alg:xrand}\;
    \If{$\beta_d < \mathbbm{U}([0,1])$ \label{alg:betad}}{    
        $x_\text{nearest}\leftarrow \text{KdTreeQuery}(x_\text{rand}, \mathcal{T}, 1)$ \label{alg:xnearest1}\;
    }
    \Else{
        $\mathcal X_\text{near} \leftarrow \text{KdTreeQuery}(x_\text{rand}, \mathcal{T}, K)$ \label{alg:kdquery}\;
            \label{alg:distStart}
            $x_\text{nearest} \leftarrow  \arg\min\limits_{x\in \mathcal X_\text{near}} \sum_i \mathcal{H}_d(o^i(x^i, x^i_\text{rand})) + \lambda_t | t - t_\text{rand} |$ \label{alg:distEnd}\;
    }
    \If{$\beta_s < \mathbbm{U}([0,1])$ \label{alg:betas}}{
        $u \leftarrow \text{ApproxSteer}(x_\text{nearest}, x_\text{rand})$ \label{alg:approxSteer}\;
    }
    \Else{
        \For{$i\in\mathcal V$ \label{alg:steerStart}}{
            $u^i \leftarrow \mathcal{H}_s(o^i(x_\text{nearest}^i, x_\text{rand}^i))$ \label{alg:steerEnd}\;
        }
    }
    $\Delta_t \leftarrow \mathbbm{U}([\underline{\Delta_t},\overline{\Delta_t}])$ \label{alg:sampledt}\;
    $x_\text{new} \leftarrow \text{Propagate}(x_\text{nearest}, u,
    \Delta_t)$ \;    
    \label{alg:propagate}
    \If{$x_\text{new} \in \ssfree$}{    
        $\mathcal{T}$.insert($x_\text{nearest},x_\text{new}$)\; 
        \If{$x_\text{new} \in \ssgoal$}{
            $\text{break}$ \label{alg:end}\; 
        }
    }
 }
\end{algorithm}

Kinodynamic RRT typically works in four major steps: i) sample a random state from a distribution (line~\ref{alg:xrand}, where $\mathbbm{U}$ denotes the uniform distribution), ii) find the closest state that is currently in the search tree (line~\ref{alg:xnearest1}), iii) steer towards the random state from the closest state in the tree (line~\ref{alg:approxSteer}), and iv) add the resulting motion to the tree if it is collision-free (lines~\ref{alg:sampledt} -- \ref{alg:end}).

Before introducing the heuristics, we define a observation model to generate local inputs for the decentralized heuristics. Given a joint state $\vs$ and a goal state $\vg^i$, the local observation for the $i$th robot is: 
\begin{align}
    \label{eq:observation}
	\vo^i &= h^i(\vs,\vg^i) = \left[ \vg^i - \vs^i, \{ \vs\ji \}_{j \in \mathcal{N}_{\mathcal{V}}^i}, \{ \vs\ji \}_{j \in \mathcal{N}_\Omega^i} \right], 
\end{align}
where $h^i$ is the observation function and the double superscript notation denotes a relative state or position, e.g., $\vs\ji = \vs^j - \vs^i$ and $\vp\ji = \vp^j - \vp^i$. In addition, $\mathcal{N}_{\mathcal{V}}^i$, $\mathcal{N}_{\Omega}^i$, denote the neighboring set of robots and obstacles, respectively. These sets are defined by the observation radius, $r_\text{sense}$, e.g.,  
\begin{equation}
	\mathcal{N}_\mathcal{V}^i = \{ j \in \mathcal{V} \ | \ \| \vp\ji \|_2 \leq r_\text{sense} \}.
\end{equation}
For obstacles, the relative state is constructed with a position to the center of the obstacle and zero velocity. 

The first heuristic is a steer function that replaces the standard \emph{ApproxSteer} function (line~\ref{alg:approxSteer}). Conventionally, \emph{ApproxSteer} samples many random control inputs, forward propagates the dynamics, and chooses the control resulting in the closest state to the desired state. Instead, our steer heuristic, $\mathcal{H}_s$, imitates the solution to a boundary value problem and maps observation to local robot control: 
\begin{align}
    u^i &= \mathcal{H}_s(o^i(x^i_\text{nearest},x^i_\text{rand})).
\end{align}
The joint-space, centralized RRT steer function is composed by calling our decentralized steer heuristic on all robots (lines \ref{alg:steerStart} -- \ref{alg:steerEnd}).
The result is that leaf node generation, $x_\text{nearest}$ is more consistent with the desired leaf generation.

The second heuristic is a distance, or cost-to-go function. For double integrator systems, a common RRT distance function is a hand-tuned weighted Euclidean distance. In contrast, our method is automatically tuned with a data-driven approach. In particular, the distance heuristic, $\mathcal{H}_d$, inputs a local observation and outputs some value corresponding to cost-to-go, where $\lambda_t$ is a scalar that balances the spatial and temporal costs: 
\begin{align}
    \tilde{c}^i &= \mathcal{H}_d(o^i(x^i_\text{nearest},x^i_\text{rand})) + \lambda_t | t_\text{nearest} - t_\text{rand} |
\end{align}
As before, the RRT distance function is composed by summing the heuristic cost for each robot (lines \ref{alg:distStart} -- \ref{alg:distEnd}). 
Thus, our distance metric accounts for obstacles and other robots. The effect on the tree growth is that the tree will choose the correct node to steer from; e.g. if a node is behind an obstacle wall, this distance function will correctly identify its large cost-to-go, even if the Euclidean distance is small. The heuristics are visualized in Fig.~\ref{fig:overview}.

Both heuristics are integrated in a probabilistic manner via hyperparameters $0 \leq \beta_d, \beta_s \leq 1$. We use the learned distance heuristic with probability $\beta_d$ and the learned steering heuristic with probability $\beta_s$. In other cases, we use the conventional functions (lines \ref{alg:betad} and \ref{alg:betas}). By adopting this scheme, similar to~\citet{DBLP:conf/icra/IchterHP18}, we maintain the probabilistic completeness guarantee of RRT.

Implementations of RRT often utilize the KD-tree data structure for the distance function~\cite{OMPL}, which reduces the complexity of the nearest-neighbor computation from $n$ to $\log n$, where $n$ is the size of the tree. However, KD-trees cannot operate on state-dependent distance metrics like $\mathcal{H}_d$. In order to maintain the same complexity as existing RRT methods, we use pre-filtering where we compute the closest $K$-neighbors with a KD-tree using a Euclidean distance, then we evaluate the distance heuristic only on the closest $K$ nodes (line \ref{alg:kdquery}). In practice, this method improves the runtime of the search significantly. We apply another common implementation technique known as goal biasing. This method acts on the sample function (line \ref{alg:xrand}) by using the goal state rather than $x_\text{rand}$ with probability $\mu$.
In practice, we found that the heuristics permitted setting the goal bias parameter to high values. 

\textit{Training}: To train the steer heuristic, $\mathcal{H}_s$, we use a similar training procedure to recent work in global-to-local learning~\cite{GLAS}. First, we collect trajectory data, from an existing centralized planner~\cite{Hoenig_2018}. Then, we transform the joint-space state-action pairs to decentralized observation-action pairs by applying a local observation model~\eqref{eq:observation} and considering the action of a single robot. Finally, we train a neural network policy with deep imitation learning on these observation-action pairs with a mean-squared loss function. We use Deep Sets~\cite{Zaheer_2017} to encode variable observation sizes in a continuous state representation. The Deep Sets architecture leverages permutation invariance of observations, resulting in a compact learning representation and relatively small model, resulting in faster inference during the tree search. The steer heuristic has the following construction: 
\begin{align}
    \label{eq:architecture}
    \mathcal{H}_s(\vo^i_t) &= \Psi([\rho_\Omega(\sum_{j \in \mathcal{N}^i_\Omega}\phi_\Omega(\vs\ji)); \rho_{\mathcal{V}}(\sum_{j \in \mathcal{N}^i_{\mathcal{V}}} \phi_\Omega(\vs\ji))]) 
\end{align}
where the semicolon denotes a stacked vector and $\Psi, \rho_\Omega, \phi_\Omega, \rho_V, \phi_V$ are feed-forward networks. 

To train the distance heuristic, $\mathcal{H}_d$, we collect trajectory data by rolling out trajectories with the steer heuristic. We transform each state to a set of local observations for each robot with~\eqref{eq:observation}. Then, we calculate the cost-to-go for each robot observation along the trajectory similar to~\eqref{eq:path_cost} by summing the control effort from the current time to the terminal time. The distance heuristic is trained in a supervised learning manner with a similar architecture as steer with a single output dimension and using the cost-to-go target instead of the action. 

\section{Experimental Results}
\label{sec:experiment_results}

\textit{Variants and Baseline}: We consider 3 variants and a baseline algorithm corresponding to the combinations of heuristics. In all figures, we label each planning solution with either: 1) BOTH that uses both heuristics, 2) DISTANCE that uses only \emph{distance} heuristic, 3) STEER that uses only \emph{steer} heuristic, and 4) NONE corresponding to plain RRT baseline. These variants are equivalently constructed by setting the value of corresponding hyperparameter, $\beta_{s,d}$, to zero.  

\textit{Experimental Setup}: We generate random $8m \times 8m$ maps with $10\%$ or $20\%$ obstacles randomly placed in a grid pattern. We consider planning problems for 4, 8, and 16 robots. The expansion time step, $\Delta_t$ is uniformly chosen from range $\underline{\Delta_t} = 0.1, \overline{\Delta_t} = 0.75$. We apply a velocity bound $\overline{v} = 0.5$ and an acceleration bound $\overline{a} = 0.5$. The robot's have radius $r_\text{robot} = 0.125$ and $r_\text{goal} = 0.2 |\mathcal{V}|$. For each different environment configuration (number of robots and obstacle density), we generate $100$ random maps, with a total time threshold of 600 seconds for each map. The distance function uses $\lambda_t = 0.05$ time bias, and, after the baseline distance function is tuned for performance, the velocities are weighed with a $0.3$ coefficient. The pre-filter step is parameterized with $K = 10$ closest nodes. The hyperparameters determining the heuristic frequency are $\beta_s = 0.5, \beta_d = 1.0$. We choose a goal bias $\mu = 0.3$ and a goal time $T = 60$. 
\begin{figure} %
    \centering
    \includegraphics[width=0.7\linewidth]{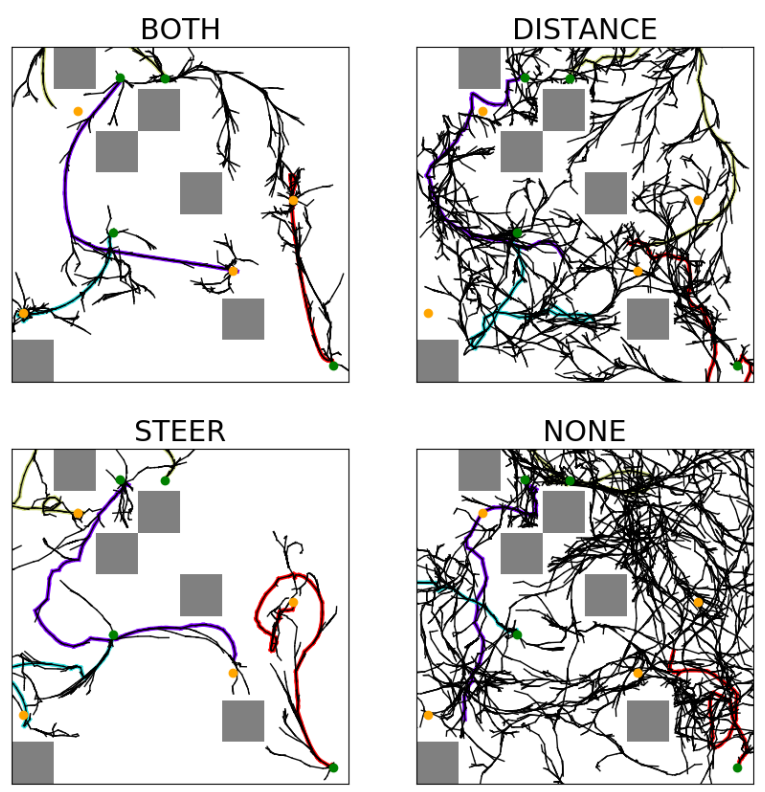}
    \caption{Four instances of joint-space planning}
    \label{js_ex}
    \vskip -0.1in
\end{figure}

\begin{figure} %
    \centering
    \includegraphics[width=0.85\linewidth]{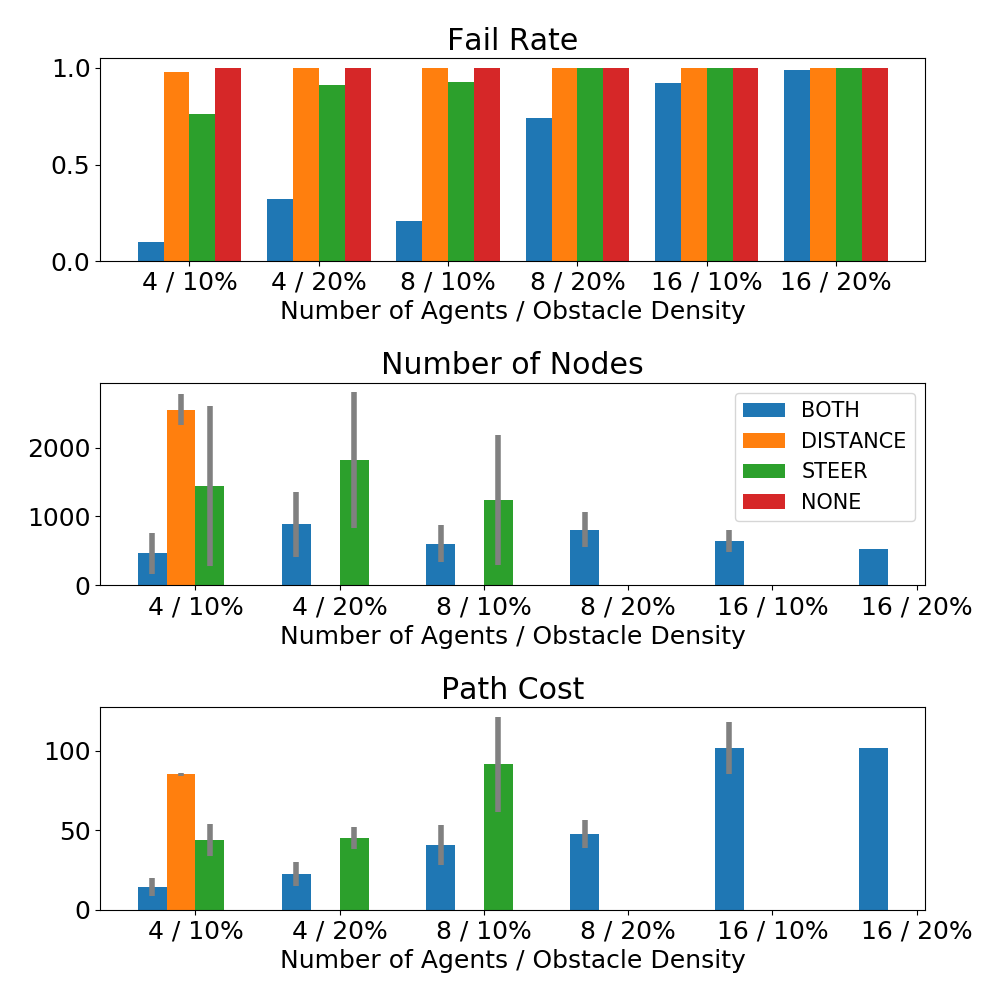}
    \vskip -0.1in
    \caption{Joint-space planner results for each problem class where each bar represents 100 instances. Missing bars indicate that the respective algorithm did not solve any instance.}
    \label{js_fig}
    \vskip -0.1in
\end{figure}

\textit{Results}: 
We generate 600 random maps for a range of robot and obstacle densities and evaluate each variant for a total of 2,400 instances and plot the collective statistics in Fig.~\ref{js_fig}. 
First, we compare the average performance across all cases of the four algorithms. The planner with both heuristics consistently has the best performance on all three metrics, empirically validating the effectiveness of both heuristics. Over the $4$ agents, $10\%$ obstacle case, the BOTH variant has an $67.7\%$ and $67.4\%$ improvement over STEER and an $81.8\%$ and $83.2\%$ improvement over DISTANCE in number of nodes and path cost, respectively. The NONE variant is unable to solve any of the cases. Next, we compare the failure rate of each of methods over problem complexity, measured through agent and obstacle density. We find that although the STEER variant seems more robust than the DISTANCE, the DISTANCE heuristic provides complementary information as the BOTH variant far outperforms the STEER variant in success rate across agent densities.

An instance of joint-space planner for our four variants is shown in Figure \ref{js_ex}. In contrast to the NONE variant that samples many states in areas near the obstacles, the DISTANCE variant correctly interprets the distance near obstacles, resulting in even sampling throughout the space. The STEER variant and BOTH variant can bias to the goal while avoiding obstacles. Combining distance and steer heuristics, BOTH variant is the most robust solution and has the least control effort and smallest curvature of trajectories.

In our supplemental material, we include implementation details, a sequential planner variant, and a swapping corridor example to demonstrate the advantages of joint-space over sequential planning. 

\section{Conclusion} 
\label{sec:conclusion}

In this work we combine decentralized learned control policies with centralized informed search. Specifically, we present two novel decentralized data-driven heuristics that enable existing sampling-based kinodynamic motion planner to find better solutions quicker, even in high-dimensional search spaces, while retaining theoretical guarantees. Unlike traditional sampling-based planners, our method can effectively plan in joint-space for up to 16 double-integrator robots, corresponding to searching through a 65-dimensional state space. 
In future work, we will investigate more robot dynamics and learning generalization to different robot density regimes. 
\bibliography{references}

\newpage

\appendix 

\section{Sequential Planner}
Sequential planners plan for one robot at a time while sequentially adding the completed robots' motions to the environment as dynamic obstacles. This decomposition finds solutions in shorter times, but is not a complete search and its performance often suffers in dense environments. We use the same parameters as the joint-space method, except the goal radius is fixed to $0.2$. 

\begin{figure}[h]
    \centering
    \includegraphics[width=0.85\linewidth]{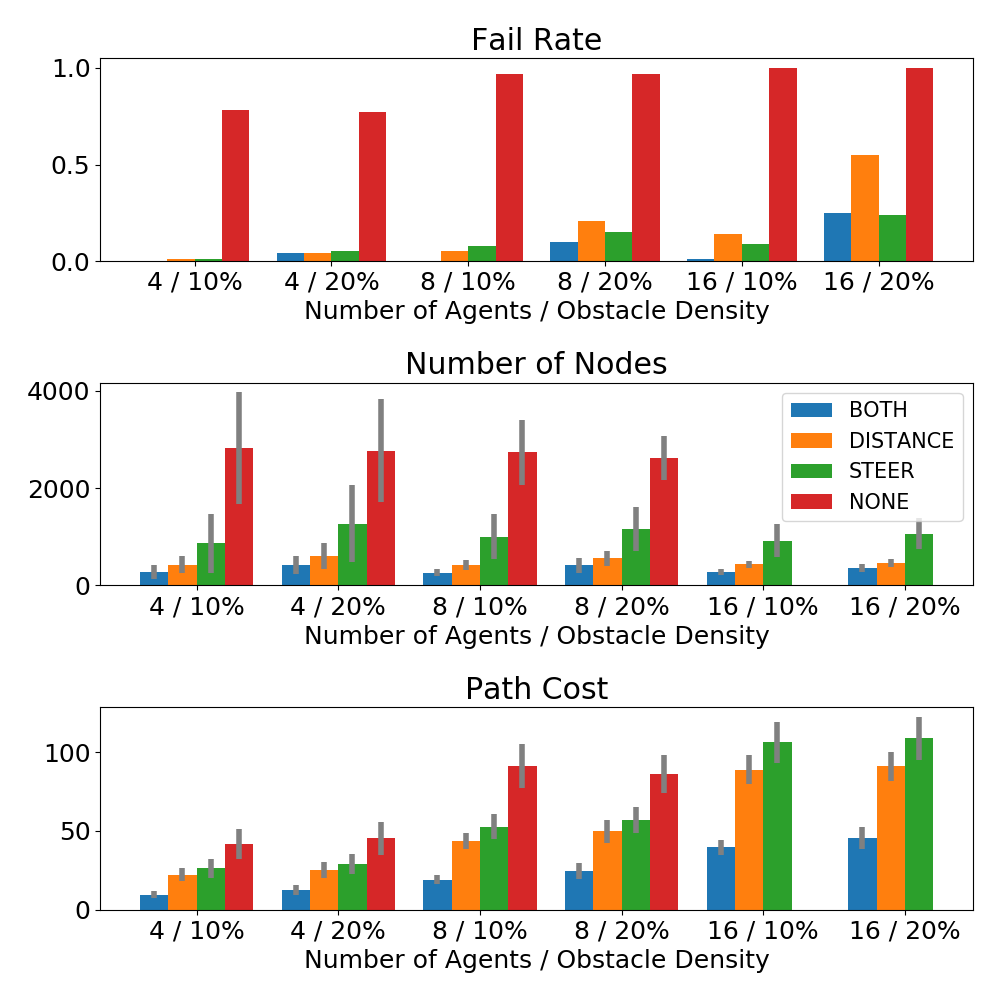}
     \caption{Sequential planner results for each problem class where each bar represents 100 instances. Missing bars indicate that the respective algorithm did not solve any instance.}
    \label{fig:sq_fig}
\end{figure}

For our numerical validation of our method, we generate and evaluate 2,400 instances motion planning problems and plot the statistics in Fig.~\ref{fig:sq_fig}.
The planner with both heuristics consistently has the best performance on all three metrics, empirically validating the effectiveness of both heuristics for a sequential planner. Our sequential planner with both heuristics improves traditional sequential sampling-based planners by producing solutions with $75\%$ lower path cost, $69\%$ lower wall-clock time, $88\%$ fewer nodes, and $93\%$ lower failure rate on average over all environments.

Next, we analyze the performance gap of the methods across complexity of cases measured through robot density. Specifically, we compare the failure rate difference, which increases from $0.78$ to $0.97$ across robot density cases from 4-robot $10\%$ obstacle density to 8-robot $10\%$ obstacle density. Moreover, the path cost ratio of NONE variant over BOTH variant also increases from $3.4$ to $4.8$. This result implies the outlook that as we demand autonomy in more complex scenarios, we will need to rely more on heuristics to search because traditional uniform sampling methods will fail to search the high-dimensional space effectively.  

\section{Swapping Example}
\label{sec:Swapping Example}
However, as sequential planning is not a complete search and its performance often suffers in dense environments, thus motivating our joint-space planning approach with better theoretical properties with respect to completeness. A classical example is the following 'swapping' problem shown in Figure \ref{sw_ex}:

\begin{figure} %
    \centering
    \includegraphics[width=0.28\linewidth]{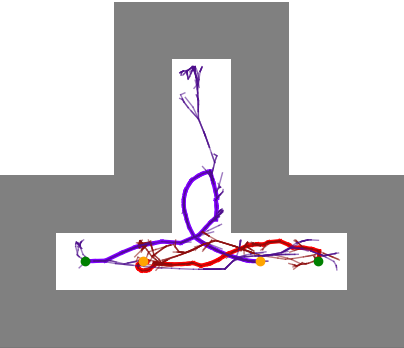}
    \includegraphics[width=0.25\linewidth]{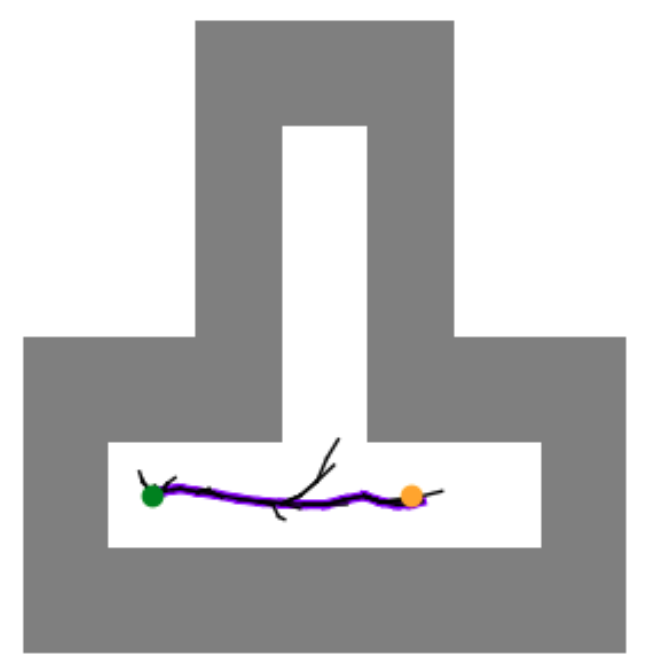}
    \includegraphics[width=0.25\linewidth]{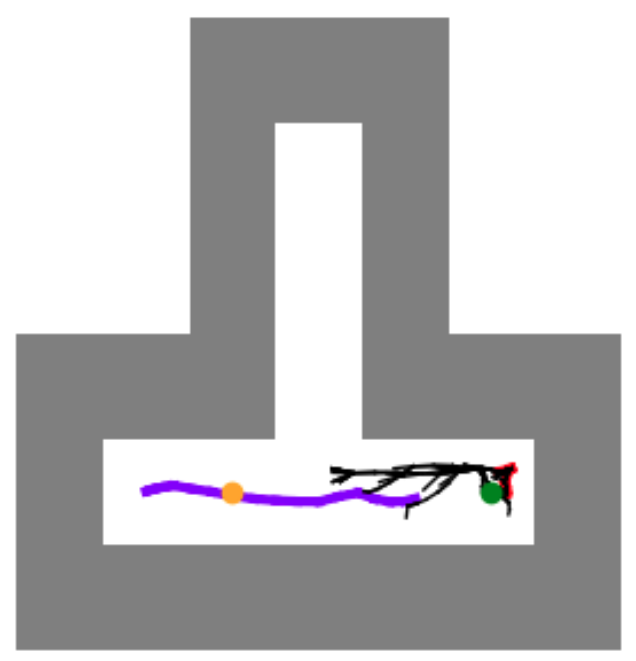}    
    \caption{From right to left: The first example shows the joint-space planner correctly finding the swapping solution. The second two examples show the sequential planner failing to solve the problem because, after it has finalized the plan for the first robot, the second robot is trapped and cannot consider the possibility of the alcove solution.}
    \label{sw_ex}
\end{figure}

In this example, the robots must swap positions through a narrow corridor, where their physical radius is such that both cannot pass at the same time. The only solution is for one robot to pass into the alcove and wait for the other to pass. We find the desired result: the joint-space planner finds the swapping solution while the sequential planner fails by nature: after it has finalized the plan for the first robot, the second robot is trapped and cannot consider the possibility of the alcove solution.

\section{Implementation Details}
\label{sec:Implementation Details}

Here we provide additional details of our learning implementation including problem and dataset generation, network architecture, and training. Our steer function is similar to the decentralized policy learned by deep imitation learning described in the GLAS~\cite{GLAS}. This method is based on an observation-action pair dataset using expert demonstration and a deep learning architecture compatible with dynamic sensing network topologies. Our distance function has a similar neural architecture. 

\textit{Problem Generation}: For data generation of our network, we use an existing implementation of a centralized global trajectory  planner~\cite{8424034} and generate $\approx 2\times10^5(200 k)$ random $8m \times 8m$ environments with $10\%$ or $20\%$ obstacles($1m \times 1m$) randomly placed in a grid pattern. We consider planning problems for 4, 8, and 16 robots. (Same as maps we used for experimental setup). The timestep for sample trajectories is set to $0.5s$ and there are $|D| = 40 \times 10^6$ data points generated in total, evenly distributed over the 6 different environment kinds. The hyperparameters of $\rho_{\mathcal{V}}$ and $\rho_\Omega$ are described in the first paragraph of experimental results.

\textit{Dataset}: As aforementioned, the expert demonstration used for our dataset is from an existing centralized planner~\cite{8424034}. This planner uses an optimization framework to minimize control effort, so the policy imitates a solution with high performance. Specifically, we create our dataset by generating fixed-size static obstacles randomly placed in a grid pattern and random start/goal positions for a variable number of robots without any collision with obstacles and each other. We use the centralized planner to compute the trajectory. For each timestep and robot, we retrieve the local observation, $\mathbf{o}^i$ by masking the non-local information, and retrieve the action $\mathbf{u}^i$ through the second derivative of the robot $i$ position. We repeat this process $n_\text{case}$ times for each robot. Our dataset $\mathcal{D}$, is:
\begin{equation}
    \mathcal{D} = \{(\mathbf{o}^i, \mathbf{u}^i) | \forall i \in \mathcal{V}, \forall k \in \{1...n_\text{case}\}, \forall t\}.
\end{equation}

\textit{Network Architecture}: The number of visible robots and obstacle for each robot greatly varies in each iteration, leading to a time-varying dimensionality of the observation. Leveraging the permutation invariance of the observation, we use Deep Set architecture~\cite{Zaheer_2017} to model variable number of robots and obstacles. In the deep set paper~\cite{Zaheer_2017}, Theorem 7 establish the following property, which informs the neural architecture of our networks: 

\textit{Theorem 1}: Let $f: [0, 1]^l \to \mathbbm{R}$ be a permutation invariant continuous function iff it has the representation:
\begin{equation}
    f(x_1, ..., x_l) = \rho\big( \sum^l_{m = 1}\phi(x_m) \big),
\end{equation}
for some continuous outer and inner function $\rho: \mathbbm{R}^{l + 1} \to \mathbbm{R}$ and $\phi: \mathbbm{R}\to\mathbbm{R}^{l + 1}$ respectively. 

Intuitively, the $\rho$ function acts to combine the contributions of each element and the $\phi$ function acts as a contribution from each element in the set. Applying Deep Sets, our steer heuristic can learn the contribution of the neighboring set of robots and obstacles with the following network structure:
\begin{align}
    \label{eq:architecture_a}
    \mathcal{H}_s(\vo^i_t) &= \Psi([\rho_\Omega(\sum_{j \in \mathcal{N}^i_\Omega}\phi_\Omega(\vs\ji)); \rho_{\mathcal{V}}(\sum_{j \in \mathcal{N}^i_{\mathcal{V}}} \phi_\Omega(\vs\ji))]),
\end{align}
where the semicolon denotes a stacked vector and $\Psi, \rho_\Omega, \phi_\Omega, \rho_V, \phi_V$ are feed-forward networks of the form:
\begin{align}
    FF(\mathbf{x}) = W^l\sigma(...W^1\sigma(\mathbf{x})),
\end{align}
where FF is a feed-forward network on input $\mathbf{x}, W^l$ is the weight matrix of the $l^\text{th}$ layer and $\sigma$ is the activation function.

The distance heuristic is trained in a supervised learning manner with similar architecture as steer with a single output dimension. The target for the distance heuristic is the cost-to-go: $c = \sum_t \|u^i_t\|_2$ summing from the time of the current state until the end of the trajectory.

For both networks, we use the mean squared loss function on the target:
\begin{align}
    \mathcal{L} = \|f(x) - y\|_2^2,
\end{align}
where $y$ is the learning target (either action from global planner or cost-to-go), and $f$ is our networks (either $\mathcal{H}_s$ or $\mathcal{H}_d$).

\textit{Learning Implementation Details}: We implement our algorithm in Python with Pytorch and our heuristics are constructed according to~\eqref{eq:architecture_a}. Both heuristics have the same number of parameters, except in the final layer where the steer outputs a 2 dimensional control vector and the distance outputs a scalar value. The $\rho_\Omega$ and $\rho_V$ networks have an input layer with 2 neurons, one hidden layer with 64 neurons, and an output layer with 16 neurons. The $\rho_\Omega$ and $\rho_V$ networks have 16 neurons in their input and output layers and one hidden layer with 64 neurons. The $\Psi$ network has an input layer with 34 neurons, one hidden layer with 64 neurons, and output layer with either one (for distance) or two (for steer) neurons. All networks use a fully connected feedforward structure with the ReLU activation function. We train our steer and distance heuristics with $3$ and $5$ million datapoints respectively, an initial learning rate of 0.001 with the PyTorch optimizer ReduceLROnPlateau function, a batch size of 32000, and train for 300 epochs.

\end{document}